\documentclass[runningheads]{llncs}
\usepackage{graphicx}
\usepackage[caption=false]{subfig}
\usepackage{amsfonts}
\usepackage{color}
\usepackage[pagebackref=false,breaklinks=true,colorlinks=true,bookmarks=false]{hyperref}
\usepackage{amsmath}
\usepackage{amssymb}
\usepackage[misc,geometry]{ifsym}
\begin{document}
\title{Font Style that Fits an Image --\\
Font Generation Based on Image Context}
\titlerunning{Font Style that Fits an Image}

\author{Taiga Miyazono\inst{1} \and
Brian Kenji Iwana\index{Iwana,Brian Kenji}\inst{1} (\Letter)\orcidID{0000-0002-5146-6818} \and
Daichi Haraguchi\inst{1} \and
Seiichi Uchida\inst{1}\orcidID{0000-0001-8592-7566}
\email{\{iwana,uchida\}@ait.kyushu-u.ac.jp}}

\authorrunning{T. Miyazono et al.}
%
\institute{Kyushu University, Fukuoka, Japan}
%
\maketitle              
\begin{abstract}
When fonts are used on documents, they are intentionally selected by designers. For example, when designing a book cover, the typography of the text is an important factor in the overall feel of the book. In addition, it needs to be an appropriate font for the rest of the book cover. Thus, we propose a method of generating a book title image based on its context within a book cover. We propose an end-to-end neural network that inputs the book cover, a target location mask, and a desired book title and outputs stylized text suitable for the cover. The proposed network uses a combination of a multi-input encoder-decoder, a text skeleton prediction network, a perception network, and an adversarial discriminator. We demonstrate that the proposed method can effectively produce desirable and appropriate book cover text through quantitative and qualitative results. The code can be found at \url{https://github.com/Taylister/FontFits}.
\end{abstract}
%
%
\section{Introduction\label{sec:intro}}
\begin{figure}[t]
    \centering
    \includegraphics[width=1.0\linewidth]{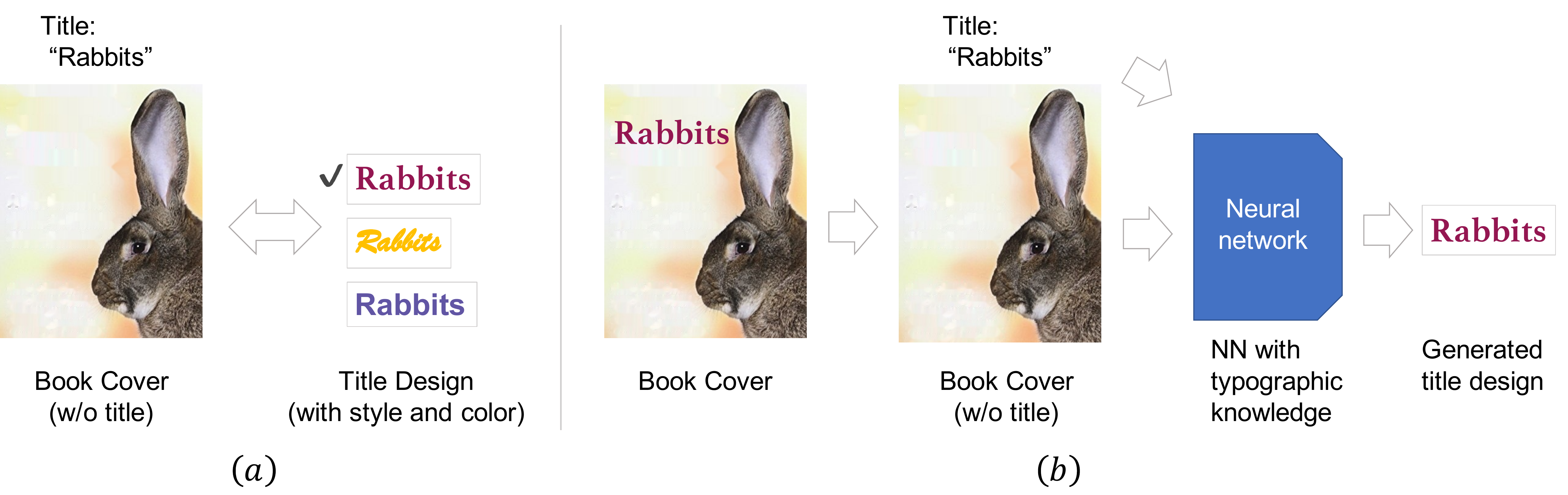}\\[-4mm]
    \caption{(a) Our question: Is there any title design that fits a book cover? (b) Our approach: If there is a specific design strategy, we can realize a neural network that learns the correlation between the cover image and the title design and then generate an appropriate title design.  \label{fig:our_task}}
    \bigskip
    \includegraphics[width=1.0\linewidth]{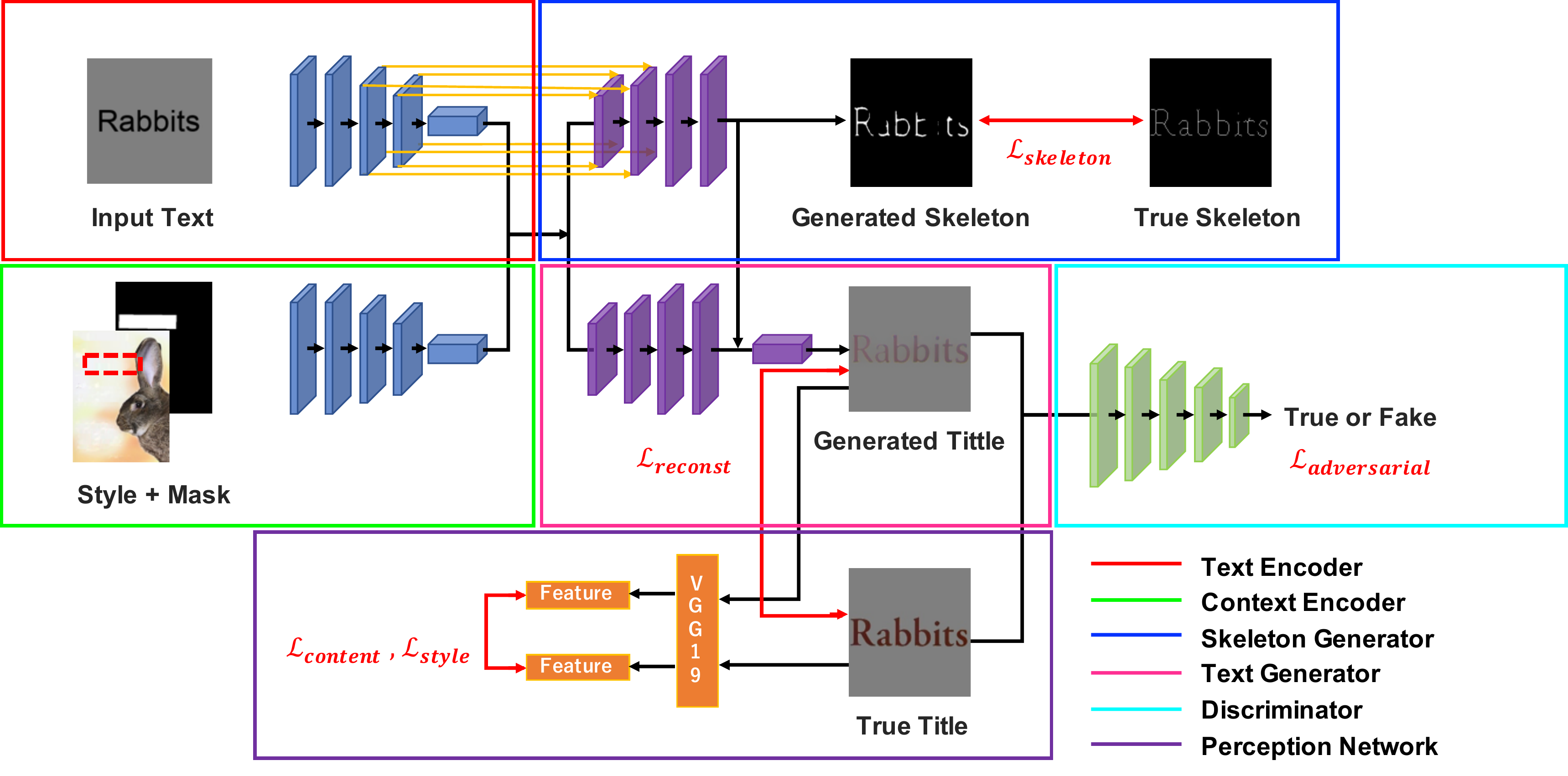}\\[-3mm]
    \caption{Overview of the proposed framework.   \label{fig:overall}}
\end{figure}

Fonts come in a wide variety of styles, and they can come with different weights, serifs, decorations, and more. 
Thus, choosing a font to use in a medium, such as book covers, advertisements, documents, and web pages, is a deliberate process by a designer.
For example, when designing a book cover, the title design (i.e., the font style and color for printing the book title) plays an important role~\cite{jubert2007,tschichold1998new}. 
An appropriate title design will depend on visual features (i.e., the appearance of the background design), as well as semantic features of the book (such as the book content, genre, and title texts).
As shown in Fig.~\ref{fig:our_task}~(a), given a background image for a book cover, typographic experts determine an appropriate title design that fits the background image. In this example, yellow will be avoided for title design to keep the visual contrast from the background; a font style that gives a hard and solid impression may be avoided by considering the impression from the cute rabbit appearance. 
These decisions about fonts are determined by the image context.

This paper aims to generate an appropriate text image for a given context image to understand their correlation.
Specifically, as shown in Fig.~\ref{fig:our_task}~(b), we attempt to generate a title image that fits the book cover image by using a neural network model trained by 104,295 of actual book cover samples. If the model can generate a similar title design to the original, the model catches the correlation inside it and shows the existence of the correlation.\par
Achieving this purpose is meaningful for two reasons. First, its achievement shows the existence of the correlation between design elements through objective analysis with a large amount of evidence. It is often difficult to catch the correlation because the visual and typographic design is performed subjectively with a huge diversity. Thus, the correlation is not only very nonlinear but also very weak. If we prove the correlation between the title design and the book cover image through our analysis, it will be an important step to understand the ``theory'' behind the typographic designs. Secondly, we can realize a design support system that can suggest an appropriate title design from a given background image. This system will be a great help to non-specialists in typographic design.
\par

In order to learn how to generate a title from a book cover image and text information, we propose an end-to-end neural network that inputs the book cover image, a target location mask, and a desired book title and outputs stylized text. As shown in Fig.~\ref{fig:overall}, the proposed network uses a combination of a Text Encoder, Context Encoder, Skeleton Generator, Text Generator, Perception Network, and Discriminator to generate the text.
The Text Encoder and Context Encoders encode the desired text and given book cover context. 
The Text Generator use skeleton-guided text generation~\cite{Wu_2019} to generate the text, and the Perception Network and adversarial Discriminator refine the results.

The main contributions of this paper are as follows:
\begin{itemize}
    \item We propose an end-to-end system of generating book title text based on the cover of a book. As far as we know, this paper presents the first attempt to generate text based on the context information of book covers.
    
    \item A novel neural network is proposed, which includes a skeleton-based multi-task and multi-input encoder-decoder, a perception network, and an adversarial network.
    
    \item Through qualitative and quantitative results, we demonstrate that the proposed method can effectively generate text suitable given the context information.

\end{itemize}

\section{Related Work\label{sec:related}}
Recently, font style transfer using neural networks~\cite{Atarsaikhan_2020} has become a growing field. 
In general, there are three approaches toward neural font style transfer, GAN-based methods, encoder-decoder methods, and NST-based methods.
A GAN-based method for font style transfer uses a conditional GAN~(cGAN). 
For example, Azadi et al.~\cite{Azadi_2018} used a stacked cGAN to generate isolated characters with a consistent style learned from a few examples. 
A similar approach was taken for Chinese characters in a radical extraction-based GAN with a multi-level discriminator~\cite{Huang_2020} and with a multitask GAN~\cite{Wu_2020}. 
Few-shot font style transfer with encoder-decoder networks have also been performed~\cite{Zhu_2020}.
Wu et al.~\cite{Wu_2019} used a multitask encoder-decoder to generate stylized text using text skeletons.
Wang et al.~\cite{Wang_2019} use an encoder-decoder network to identify text decorations for style transfer.
In Lyu et al.~\cite{Lyu_2017} an autoencoder guided GAN was used to generate isolated Chinese characters with a given style. 
There also have been a few attempts at using NST~\cite{Gatys_2016} to perform font style transfer between text~\cite{Atarsaikhan_2020,Gomez_2019,Narusawa_2019,Ter_Sarkisov_2020}.

An alternative to text-to-text neural font style transfer, there have been attempts to transfer styles from arbitrary images and patterns. 
For example, Atarsaikhan et al.~\cite{Atarsaikhan_2020} proposed using a distance-based loss function to transfer patterns to regions localized to text regions. 
There are also a few works that use context information to generate text. 
Zhao et al.~\cite{Zhao_2018} predicted fonts for web pages using a combination of attributes, HTML tags, and images of the web pages. 
This is similar to the proposed method in that the context of the text is used to generate the text. 
Yang et al.~\cite{yang2020learning} stylized synthetic text to become realistic scene text.
The difference between the proposed method and these methods is that we propose inferring the font style based on the contents of the desired medium and use it in an end-to-end model to generate text.

\section{Automatic Title Image Generation}

The purpose of this study is to generate an image of the title text with a suitable font and color for a book cover image. 
Fig.~\ref{fig:overall} shows the overall architecture of the proposed method.
The network consists of 6 modules:  Text Encoder, Context Encoder, Text Generator, Skeleton Generator, Perception Network, and Discriminator. 
The Text Encoder and Context Encoder extracts text and styles from the input text and style cover. 
The Generator generates the stylized text suitable for the book cover input, and the Skeleton Generator creates a text skeleton to guide the Generator.
The Perception Network and Discriminator help refine the output of the Text Generator.

\subsection{Text Encoder}
\begin{figure}[t]
    \centering
    \includegraphics[width=1.0\linewidth]{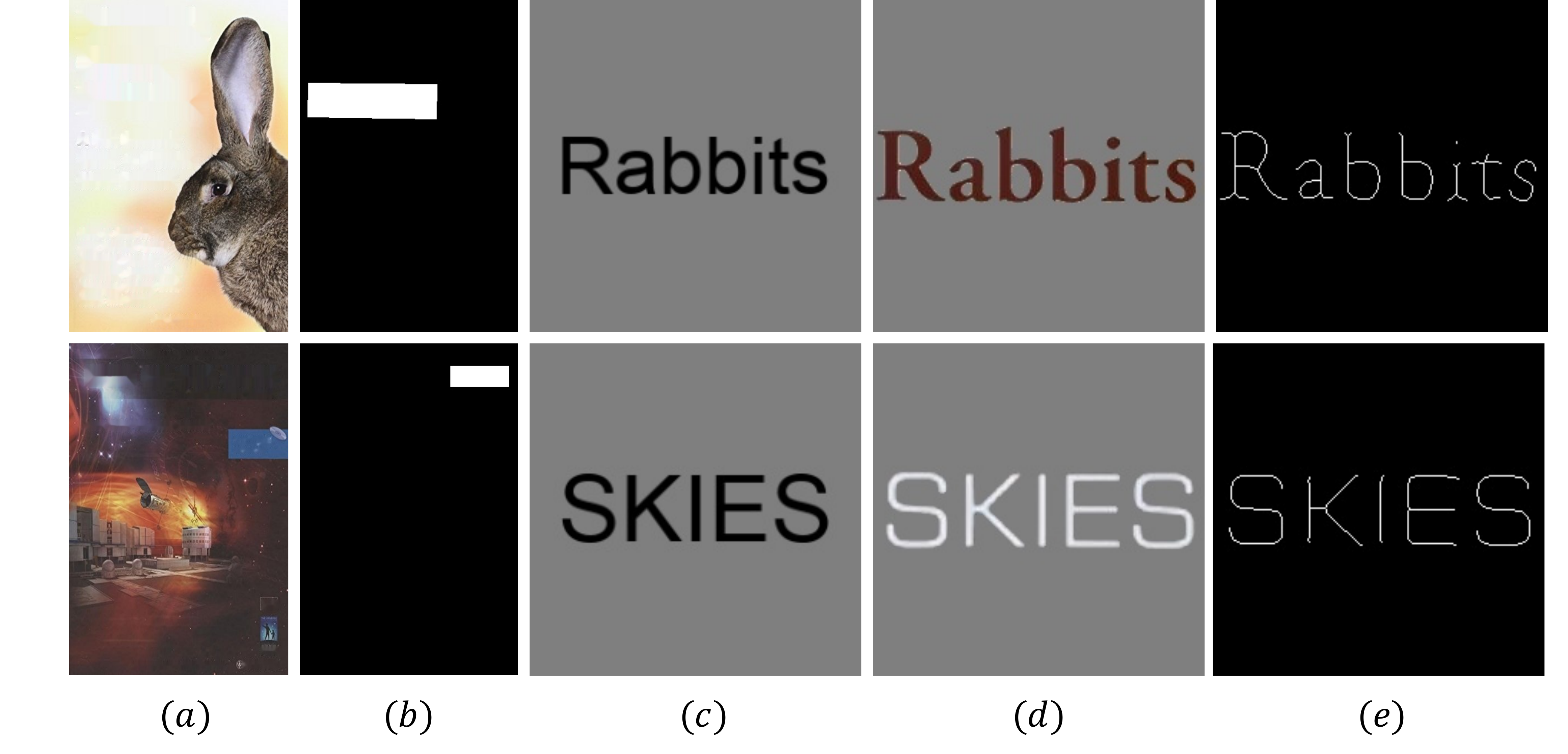}
    \caption{Training data example. From left to right: style cover image, style cover mask image, input text image, true title image, true skeleton image.}
    \label{fig:traningdata-example}
\end{figure}

The Text Encoder module extracts character shape features from an image $I_{t}$ of the input text. 
These features are used by the Text Generator and the Skeleton Generator to generate their respective tasks.
As shown in Fig.~\ref{fig:traningdata-example}, the Text Encoder input is an image of the target text rendered on a background with a fixed pixel value. 

The encoder consists of 4 convolutional blocks with residual connections~\cite{He2015}. 
The first block consists of two layers of 3$\times$3 convolutions with stride 1. 
The subsequent blocks contain a 3$\times$3 convolutional layers with stride 2 and two convolutional layers with stride 1.
Leaky Rectified Linear Units~(Leaky ReLU)~\cite{maas_2013} are used as the activation function for each layer. The negative slope is set to 0.2.
There are skip-connections between second and third convolutional blocks and the convolutional blocks of the Skeleton Generator, which is described later.

\subsection{Context Encoder}
The Context Encoder module extracts the style features from the cover image $S_{c}$ and the location mask image $S_{m}$. 
Examples of the covers $S_{c}$ and the location mask $S_{m}$ are shown in Fig.~\ref{fig:traningdata-example}.
The cover image $S_{c}$ provides the information about the book cover, such as color, objects, layout, etc.) for the Context Encoder to predict the style of the font and the location mask $S_{m}$ provides target location information. 
It should be noted that the cover image $S_{c}$ has the text inpainted, i.e., removed. 
Thus, the style is inferred solely based on the cover and not on textual cues.

The Context Encoder input is constructed of $S_{c}$ and $S_{m}$ concatenated in the channel dimension. 
Also, the Context Encoder structure is the same as the Text Encoder, except that the input is only 2 channels (as opposed to 3 channels, RGB, for the Text Encoder).
As the input of the generators, the output of the Context Encoder is concatenated in the channel dimension with the output of the Text Encoder.

\subsection{Skeleton Generator\label{seq:Skeleton}}
In order to improve the legibility of the generated text, a skeleton of the input text is generated and used to guide the generated text. 
This idea is called the Skeleton-guided Learning Mechanism~\cite{Wu_2019}. 
This module generates a skeleton map, which is the character skeleton information of the generated title image. 
As described later, the generated skeleton map is merged into the Text Generator. 
By doing so, the output of the Text Generator is guided by the skeleton to generate a more robust shape.

This module is an upsampling CNN that uses four blocks of convolutional layers. 
The first block contains two standard convolutional layers, and the three subsequent blocks have one transposed convolutional layer followed by two standard convolutional layers. 
The transposed convolutional layers have 3$\times$3  transposed convolutions at stride 2, and the standard convolutional layers have 3$\times$3 convolutions at stride 1.
All the layers use Batch Norm and LeakyReLU.
In addition, there are skip-connections between the Text Encoder and after the first layer of the second and third convolutional blocks.

To train the Skeleton Generator, the following skeleton loss $\mathcal{L}_{\mathrm{skeleton}}$ is used to train the combined network:
\begin{equation}
  \mathcal{L}_{\mathrm{skeleton}} = 1 - \frac{2\sum_{i}^{N} (T_{\mathrm{skeleton}})_{i=1}(O_{\mathrm{skeleton}})_{i}}{\sum_{i=1}^{N} (T_{\mathrm{skeleton}})_{i} + \sum_{i=1}^{N} (O_{\mathrm{skeleton}})_{i}},
\end{equation}
where $N$ is the number of pixels, $T_{\mathrm{skeleton}}$ is true skeleton map, and $O_{\mathrm{skeleton}}$ is the output of Skeleton Generator. This skeleton loss is designed based on DiceLoss~\cite{Milletari_2016}. 

\subsection{Text Generator\label{seq:Skeleton}}
The Text Generator module takes the features extracted by Text Encoder and Style Encoder and outputs an image of the stylized title. 
This output is the desired result of the proposed method. 
It is an image of the desired text with the style inferred from the book cover image.

The Text Generator has the same structure as the Skeleton Generator, except with no skip-connections and with an additional convolutional block.
The additional convolutional block combines the features from the Text Generator and the Skeleton Generator. 
As described previously, the Skeleton Generator generates a skeleton of the desired stylized text. To incorporate the generated skeleton into the generated text, the output of the Skeleton Generator is concatenated with the output of the fourth block of the Text Generator.
The merged output is further processed through a $3\times3$ convolutional layer with a stride of 1, Batch Normalization, and Leaky ReLU activation.
The output $O_{t}$ of the Text Generator has a tanh activation function. 

The Text Generator is trained using a reconstruction loss $\mathcal{L}_{\mathrm{reconst}}$. 
The reconstruction loss is Mean Absolute Error between the generated output $O_{t}$ and the ground truth title text $T_{t}$, or:
\begin{equation}
  \mathcal{L}_{\mathrm{reconst}} =|T_{t} - O_{t}|.
\end{equation}
While loss guides the generated text to be similar to the original text, it is only one part of the total loss.
Thus, the output of the Text Generator is not strictly the same as the original text.

\subsection{Perception Network}
To refine the results of the Text Generator, we use a Perception Network. 
Specifically, the Perception Network is used to increase the perception of the generated images~\cite{johnson2016perceptual,Wu_2019}. 
To do this, the output of the Text Generator is provided to a Perception Network, and two loss functions are added to the training. 
These loss functions are the Content Perceptual loss $\mathcal{L}_{\mathrm{content}}$ and the Style Perceptual loss $\mathcal{L}_{\mathrm{style}}$. 
The Perception Network is a VGG19~\cite{simonyan2014very} that was pre-trained on ImageNet~\cite{Jia_Deng_2009}.

Each loss function compares differences in the content and style of the features extracted from the Perception Network when provided the generated title and the ground truth title images.
The Content Perceptual loss minimizes the distance between the extracted features of the generated title images and the original images, or:
\begin{eqnarray}
  \mathcal{L}_{\mathrm{content}} &=&\sum_{l \in \mathcal{F}}|P_{l}(T_{t})-P_{l}(O_{t})|,
\end{eqnarray}
where $P$ is the Perception Network, $P_{l}(\cdot)$ is the feature map of $P$'s $l$-th layer given the generated input image $O_t$ or ground truth image $T_t$, and $\mathcal{F}$ is the set of layers used in these loss.
In this case, $\mathcal{F}$ is set to the relu1\_1, relu2\_1, relu3\_1, relu4\_1, and relu5\_1 layers of VGG19. 
The Style Perceptual loss compares the texture and local features extracted by the Perception Network, or:
\begin{eqnarray}
  \mathcal{L}_{\mathrm{style}} &=&\sum_{l \in \mathcal{F}}|\Psi^{P}_{l}(T_{t})-\Psi^{P}_{l}(O_{t})|,
\end{eqnarray}
where $\Psi^{P}_{l}(\cdot)$ is a Gram Matrix, which has $C_{l} \times C_{l}$ elements. 
Given input $I\in\{O_t, T_t\}$, $P_{l}(I)$ has a feature map of shape $C_{l} \times H_{l} \times W_{l}$. 
The elements of Gram Matrix $\Psi^{P}_{l}(I)$ are given by:
\begin{eqnarray}
  \Psi^{P}_{l}(I)_{c,c^{'}} &=&\frac{1}{C_{l}H_{l}W_{l}} \sum_{h  = 1}^{H_{l}} \sum_{w = 1}^{W_{l}} P_{l}(I)_{h,w,c}P_{l}(I)_{h,w,c^{'}},
\end{eqnarray}
where $c$ and $c'$ are each element of $C_t$. 
By minimizing the distance between the Gram Matrices, a style consistency is enforced. 
In other words, the local features of both images should be similar.

\subsection{Discriminator} 
In addition to the Perception Network, we also propose the use of an adversarial loss $\mathcal{L}_{\mathrm{adversarial}}$ to ensure that the generated results are realistic. 
To use the adversarial loss, we introduce a Discriminator to the network architecture. 
The Discriminator distinguishes between whether the input is a real title image or a fake image. 
In this case, we use the true tile image $T_t$ as the real image and the Text Generator's output $O_t$ as the fake image.

The Discriminator in the proposed method follows the structure of the DCGAN \cite{Li_2018}.
The Discriminator input goes through 5 down-sampling 5$\times$5 convolutional layers with stride 2 and finally a fully-connected layer. 
The output is the probability that the input is a true title image. Except for the output layer, the LeakyReLU function is used. 
At the output layer, a sigmoid function is adopted. The following adversarial loss is used to optimize the entire generation model and the Discriminator:
\begin{equation}
  \mathcal{L}_{\mathrm{adversarial}} = \mathbb{E}_{I_{t},T_{t}}[\log{D(I_{t},T_{t})}] + \mathbb{E}_{I_{t},S}[\log\{1 - D(I_{t},G(I_{t},S))\}],
\end{equation}
where $G$ is the whole generation module, $D$ is the discriminator, $S$ is the style condition$(S_c, S_m)$, $T_{t}$ is the true title image, and $I_{t}$ is the input title text. 

\subsection{Title image generation model}
As we have explained, the proposed network can generate title text suitable for a style. 
As a whole, the Text Encoder receives an input text image $I_{t}$ and the Context Encoder receives a style image $(S_c,S_m)$, and the Skeleton Generator outputs a skeleton map $O_{\mathrm{skeleton}}$ and the Text Generator outputs a title image $O_{t}$. 
This process is shown in Fig.~\ref{fig:overall}.
The training is done with alternating adversarial training with the Discriminator and the rest of the modules in an end-to-end manner. 
The Text Generator, Skeleton Generator, Text Encoder, and Context Encoder are trained using a total loss $\mathcal{L}_\mathrm{total}$ through the Text Generator, where 
\begin{equation}
  \mathcal{L}_{\mathrm{total}} = w_{1}\mathcal{L}_{\mathrm{reconst}} + w_{2}\mathcal{L}_{\mathrm{skeleton}} + w_{3}\mathcal{L}_{\mathrm{adversarial}} + w_{4}\mathcal{L}_{\mathrm{content}} + w_{5}\mathcal{L}_{\mathrm{style}}.
  \label{eq:min_max}
\end{equation}
Variables $w_{1}$ to $w_{5}$ are weights for each of the losses. 

\section{Experimental setup}

\subsection{Dataset}
In the experiment, as shown in Fig.~\ref{fig:traningdata-example}, to train the network, we need a combination of the full book cover image without text, a target location mask, a target plain input text, the original font, and the skeleton of the original words. 
Thus, we use a combination of the following datasets and pre-processing to construct the ground truth. 

We used the Book Cover Dataset~\cite{iwana2016judging}. 
This dataset consists of 207,572 book cover images\footnote{\url{https://github.com/uchidalab/book-dataset}}. 
The size of the book cover image varies depending on the book cover, but for this experiment, we resize the images to 256$\times$256 pixels in RGB.
\begin{figure}[t]
\centering
 \subfloat[Detected Text]{ \label{fig:title}
  \includegraphics[width=0.25\columnwidth]{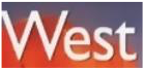}
 }
 \subfloat[Text Mask]{ \label{fig:title_mask}
  \includegraphics[width=0.25\columnwidth]{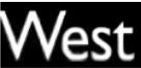}
 }
 \subfloat[Extracted Text]{ \label{fig:ex_title}
  \includegraphics[width=0.25\columnwidth]{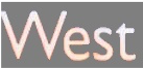}
 }
 \caption{Mask generation and title extraction.}
 \label{CSnet-result}
\end{figure}

To ensure that the generated style is only inferred by the book cover and not any existing text, we remove all of the text from the book covers before using them. 
To remove the text, we first use Character Region Awareness for Text Detection~(CRAFT)~\cite{Baek_2019_Detection} to detect the text, then cut out regions defined by the detected bounding-boxes with dilated with a 3$\times$3 kernel.
CRAFT is an effective text detection method that uses a U-Net-like structure to predict character proposals and uses the affinity between the detected characters to generate word-level bounding-box proposals. 
Then, Telea's inpainting~\cite{telea2004image} is used to fill the removed text regions with the plausible background area. 
The result is images of book covers without the text.

For the title text, CRAFT is also used. 
The text regions found by CRAFT are recognized using the scene text recognition method that was proposed by Baek et al.~\cite{Baek_2019_STRcomparisons}. 
Using the text recognition method, we can extract and compare the text regions to the ground truth. 
Once the title text is found, we need to extract the text without the background, as shown in Fig.~\ref{CSnet-result}. 
We generate a text mask (Fig.~\ref{fig:title_mask}) using a character stroke separation network to perform this extraction. 
The character stroke separation network is based on pix2pix~,\cite{isola2017image} and it is trained to generate the text mask based on the cropped detected text region of the title. 
By applying the text mask to the detected text, we can extract the title text without a background. 
The background is replaced with a background with pixel values (127.5, 127.5, 127.5). 
Since the inputs are normalized between [-1, 1], the background represents a ``0'' value.
Moreover, a target location mask (Fig.~\ref{fig:traningdata-example}b) is generated by masking the bounding-box of the title text.
The plain text input is generated using the same text but in the font ``Arial.'' 
Finally, the skeleton of the text is obtained using the skeletonization method of Zhang et al.~\cite{zhang1984fast}. 

\begin{figure}[t]
    \centering
    \includegraphics[width=1.0\linewidth]{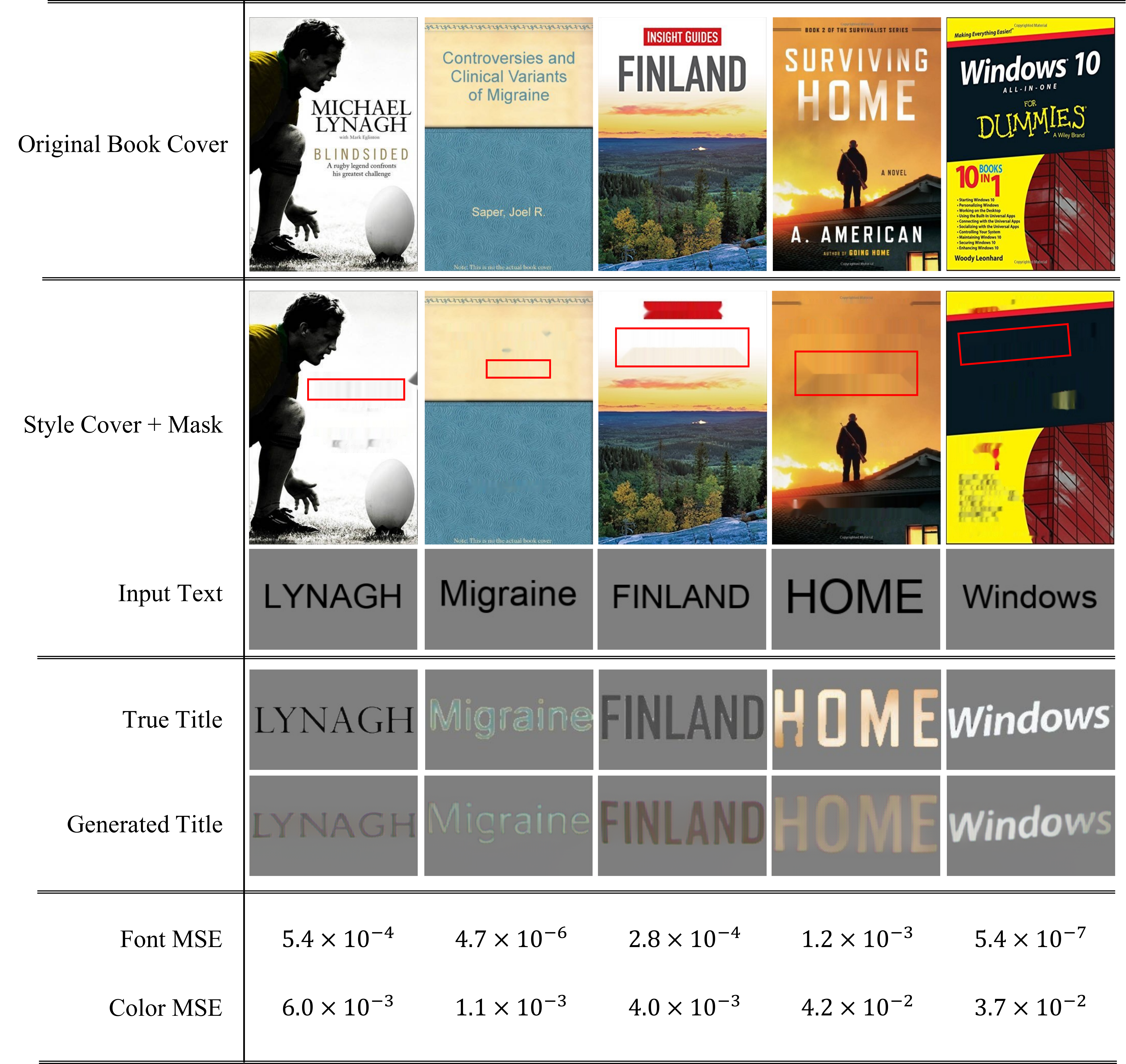}\\[-3mm]
    \caption{Successful generation results. }
    \label{fig:good}
\end{figure}
\subsection{Implementation details}

In this experiment, for training the proposed model on high-quality data pairs, only images where the character recognition results~\cite{Baek_2019_STRcomparisons} of the region detected by CRAFT (Fig.~\ref{fig:title}) and the image created by the character stroke separation network (Fig.~\ref{fig:ex_title}) match were used. As a result, our proposed model has been trained on 195,922 title text images and a corresponding 104,925 book cover images.
Also, 3,702 title text images and 1,000 accompanying book cover images were used for the evaluation. 
The training was done end-to-end using batch size 8 for 300,000 iterations. 
We used Adam \cite{kingma2014adam} as optimizer, and set the learning coefficient $lr = 2\times 10^{-4}$, $\beta_{1}=0.5$, and $\beta_{2}=0.999$. We also set $w_{1}=1$, $w_{2}=1$, $w_{3}=1.0 \times 10^{-2}$, $w_{4}=1$, and $w_{5}=1.0 \times 10^{3}$ for the weighting factors of the losses.
The weights are set so that the scale of each loss value is similar.
For $w_{3}$, we set a smaller value so that Discriminator does not have too much influence on the generation module's training and can generate natural and clear images~\cite{ledig2016photorealistic}.


\subsection{Evaluation metrics}

For quantitative evaluation of the generate title images, we use standard metrics and non-standard metrics. For the standard metrics, 
Mean Absolute Error (MAE), Peak Signal-to-Noise Ratio (PSNR), and Structural Similarity (SSIM)~\cite{wang2004image} are used.
We introduce two non-standard metrics for our title generation task, Font-vector Mean Square Error (Font MSE) and Color Mean Square Error (Color MSE). 
Font MSE evaluates the MSE between font style vectors of the original (i.e., ground-truth) and the generated title images. The font style vector is a six-dimensional vector of the likelihood of six font styles: serif, sans, hybrid, script, historical, and fancy. 
The font style vector is estimated by a CNN trained with text images generated by SynthText~\cite{Gupta16} with 1811 different fonts.
Color MSE evaluates the MSE between three-dimensional RGB color vectors of the original and the generated title images. The color vector is given as the average color of the stroke detected by the character stroke separation network.
These evaluations are only used in the experiment where the target text is the same as the original text. 
It should be noted that the three standard evaluations MAE, PSNR, and SSIM, can only be used when the target text is the same as the original text. 
However, we can use Font MSE and Color MSE, even when the generated text is different because they measure qualities that are common to the generated text and the ground truth text.

\begin{figure}[t]
    \centering
    \includegraphics[width=1.0\linewidth]{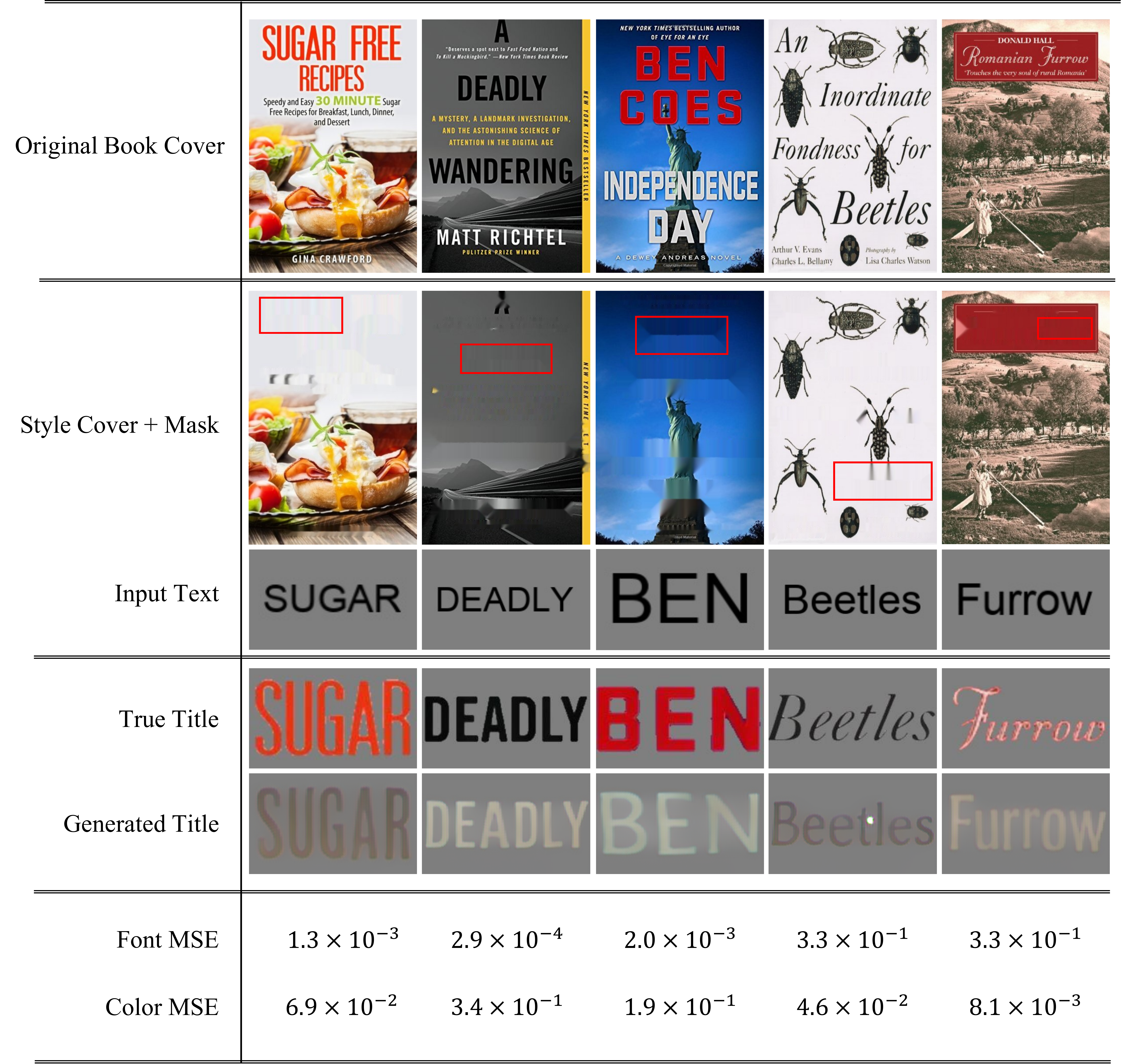}
    \caption{Failure results.}
    \label{fig:bad}
\end{figure}

\begin{figure}[t]
    \centering
    \includegraphics[width=1.0\linewidth]{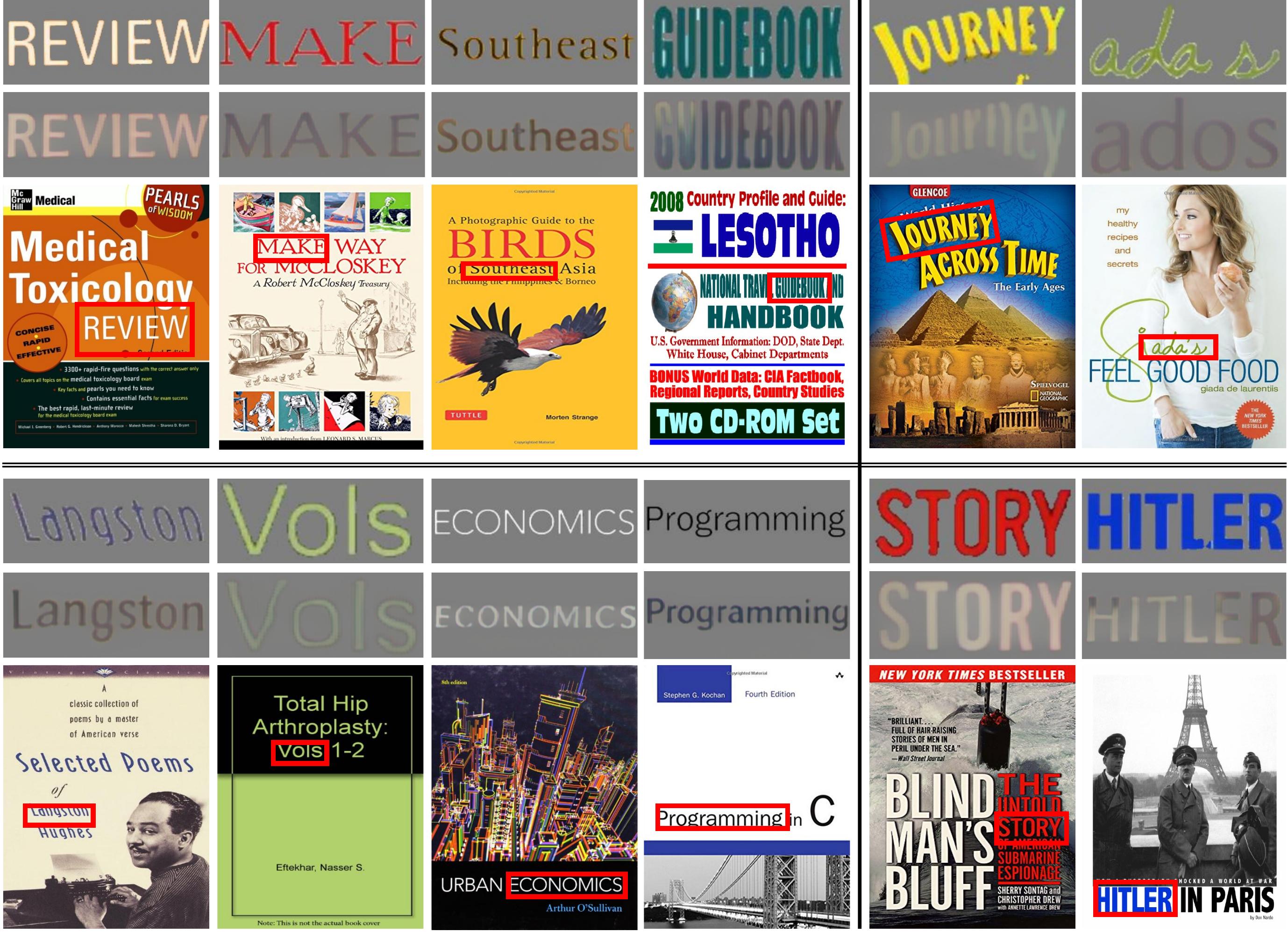}
    \caption{Other generation results. The top and bottom rows show the results selected by FontMSEs and ColorMSEs, respectively. The left four columns show lower MSE (i.e., successful) cases and the remaining two show the higher MSE (i.e., failure) cases. Each result is comprised of the true (top) and generated (middle) titles and the whole book cover image (bottom).}
    \label{fig:additional}
\end{figure}
\section{Experimental results\label{sec:results}}
\subsection{Qualitative evaluation}\label{sec:results}

This section discusses the relationship between the quality of the generated images and the book cover images by showing various samples generated by the network and the corresponding original book cover images. 
Fig.~\ref{fig:good} shows the samples where the network generates a title image close to the ground truth successfully, that is, with smaller Font MSE and Color MSE. From the figure, we can see that the strokes, ornaments, and the size of the text are reproduced. Especially, the first example shows the serif font is also reproducible even if the input text is always given as a sans-serif image.  
\par
Among the results in Fig.~\ref{fig:good}, the ``Windows'' example clearly demonstrates that the proposed method can predict the font and style of the text given the book cover and the target location mask. 
This is due to the book cover being a recognizable template from the ``For Dummies'' series in which other books with similar templates exist in the training data. 
The figure demonstrates that the proposed method effectively infers the font style from the book cover image based on context clues alone.

Fig.~\ref{fig:bad} shows examples where the generator could not successfully generate a title image close to the ground truth. The first, second, and third images in Fig.~\ref{fig:bad} show examples of poor coloring in particular. 
The fourth and fifth images show examples where the proposed method could not predict the font shape.
For the ``SUGAR'' and ``BEN'' images, there are no clues that the color should be red.
In the ``DEADLY'' book cover image, one would expect light text on the dark background. However, the original book cover used dark text on a dark background. 
For the ``Beetles'' and ``Furrow'' examples, the fonts are highly stylized and difficult to predict. \par

Fig.~\ref{fig:additional} shows several additional results including successful and failure cases. Even in this difficult estimation task from a weak context, the proposed method gives a reasonable style for the title image. The serif for ``Make'' and the thick style for ``GUIDEBOOK'' are promising. We also observe that peculiar styles, such as very decorated fonts and vivid colors, are often difficult to recover from the weak context.
\par
\begin{figure}[t]
    \centering
    \includegraphics[width=1.0\linewidth]{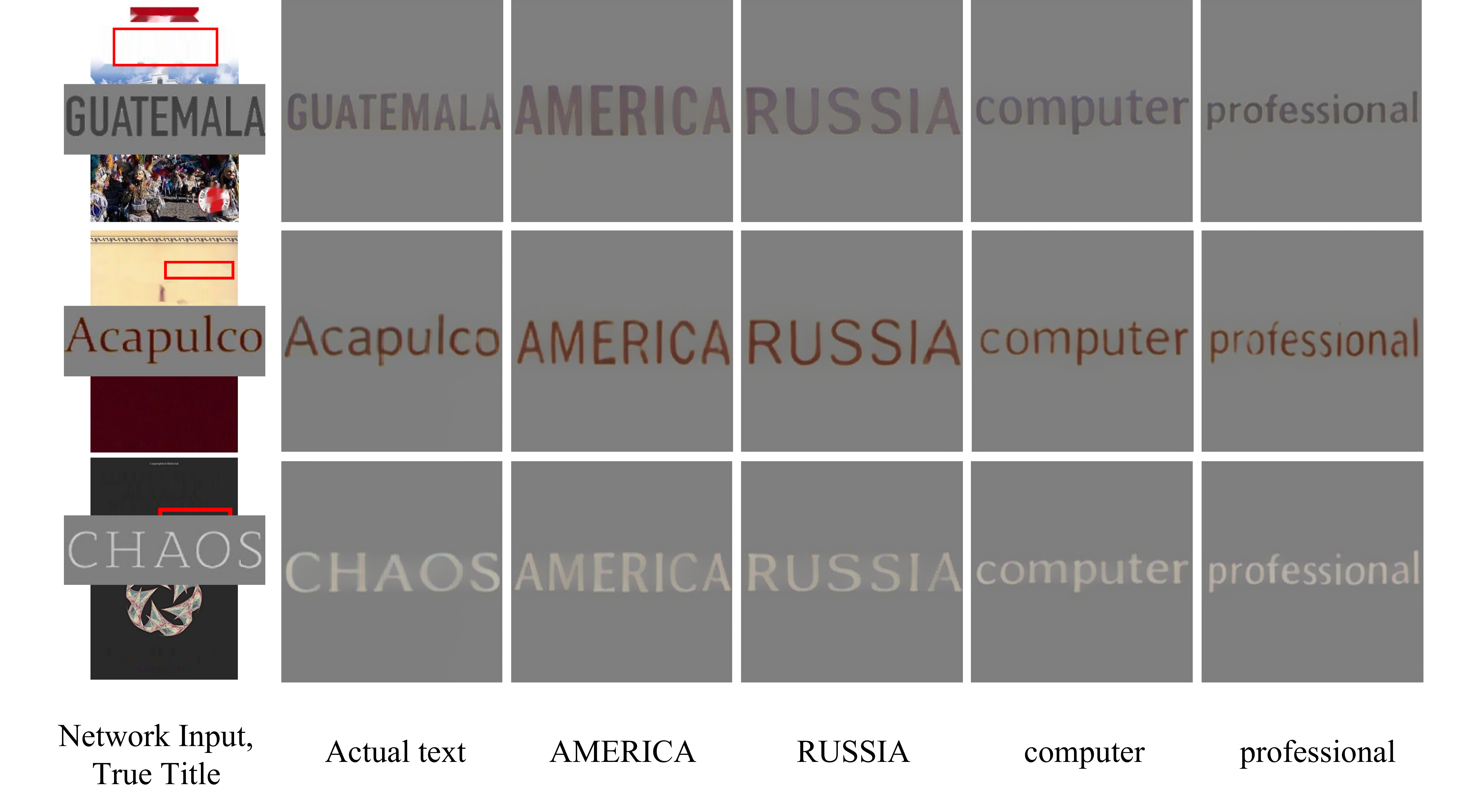}
    \caption{Generation results using text that is different from the original text.}
    \label{fig:diff_text}
\end{figure}

Finally, in Fig.~\ref{fig:diff_text}, we show results of using the proposed method, but with text that is different from the ground truth. 
This figure demonstrates that we can produce any text in the predicted style. 

\subsection{Ablation study}
To measure the importance of the components of the proposed method, quantitative and qualitative ablation studies are performed. The network of the proposed method consists of six modules and their associated loss functions. Therefore, we measure the effects of the components. All experiments are performed with the same settings and the same training data. 

The following evaluations are performed:
\begin{itemize}
    \item \textbf{Proposed}: The evaluation with the entire proposed model.
    \item \textbf{Baseline}: Training is performed only using the Text Encoder and Text Generator with the reconstruction loss $\mathcal{L}_\mathrm{reconst}$.
    \item \textbf{w/o Context Encoder}: The proposed method but without the Context Encoder. The results are expected to be poor because there is no information about the style to learn from.
    \item \textbf{w/o Skeleton Generator}: The proposed method but with no guidance from the Skeleton Generator and without the use of the skeleton loss $\mathcal{L}_\mathrm{skeleton}$.
    \item \textbf{w/o Discriminator}: The proposed method but without the Discriminator and the adversarial loss $\mathcal{L}_\mathrm{adversarial}$.
    \item \textbf{w/o Perception Network}: The proposed method but without the Perception Network and the associated losses  $\mathcal{L}_\mathrm{content}$ and $\mathcal{L}_\mathrm{style}$.
\end{itemize}

\begin{table}[t]
\centering
\caption{Quantitative evaluation results.}
\begin{tabular}{l|l|l|l|l|l}
\hline
Method                   & \multicolumn{1}{c|}{MAE $\downarrow$} & \multicolumn{1}{c|}{PSNR $\uparrow$} & \multicolumn{1}{c|}{SSIM $\uparrow$} & \multicolumn{1}{c|}{Font MSE$ \downarrow$}
& \multicolumn{1}{c}{Color MSE$ \downarrow$} \\ \hline
Baseline                 & 0.041 & 20.49 & 0.870 & 0.174 & 0.102  \\ \hline
Proposed     & {\bf 0.035} & {\bf 21.58} & {\bf 0.876} & {\bf 0.062}  & 0.064  \\ 
 \  w/o Context Encoder    & 0.036 & 20.79 & 0.872 & 0.126 & 0.093   \\ 
 \  w/o Discriminator      & {\bf 0.035} & 21.46 & 0.814 & 0.105 & 0.081  \\ 
 \  w/o Perception Network      & 0.061 & 19.39 & 0.874 & 0.085 & {\bf 0.058}   \\ 
 \  w/o Skeleton Generator & {\bf 0.035} & 21.09 & 0.874 & 0.112 & 0.080    \\ \hline
\end{tabular}
\label{tb:ablation}
\end{table}

\begin{figure}[t]
    \centering
    \includegraphics[width=1.0\linewidth]{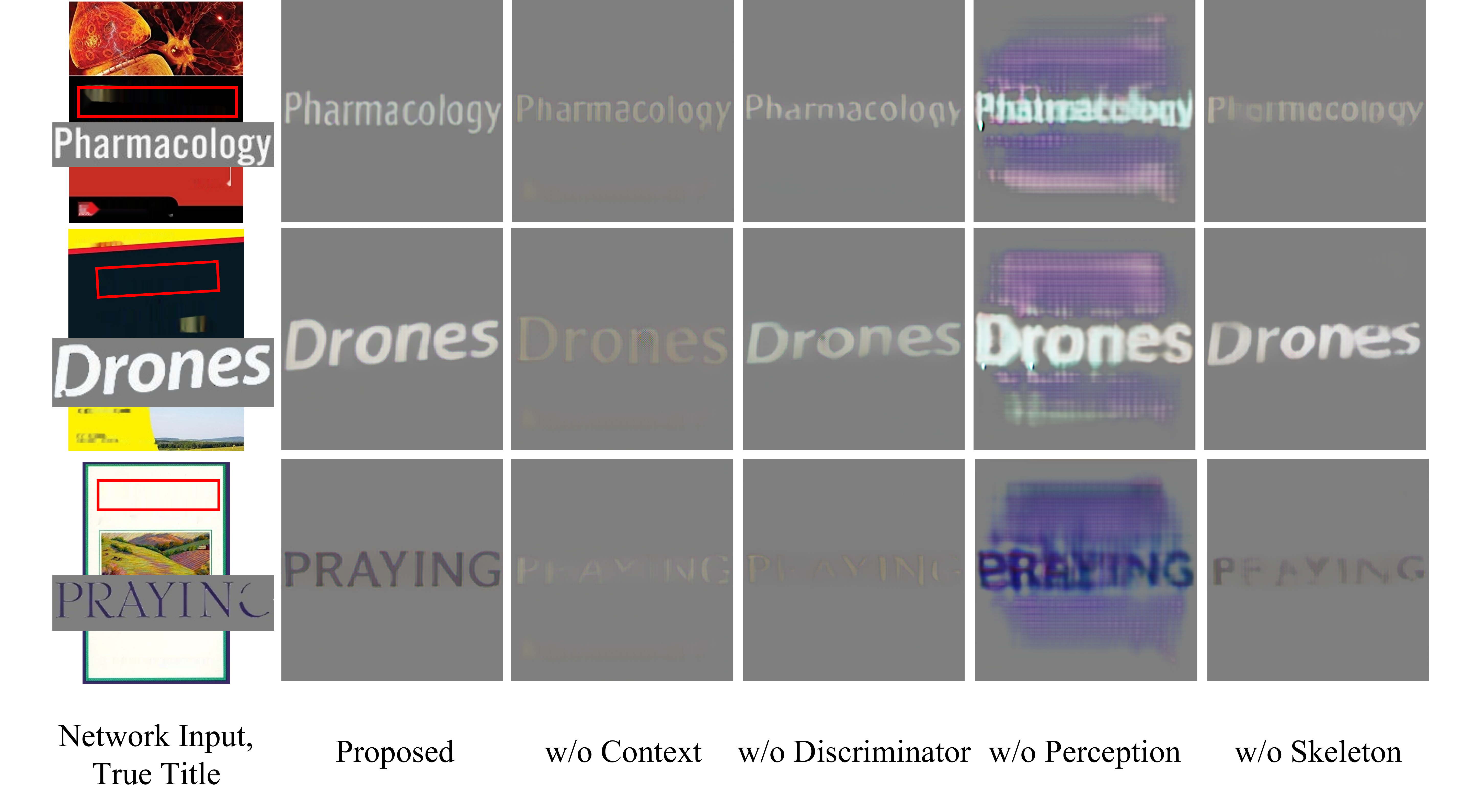}
    \caption{Results of the ablation study. } 
    \label{fig:ablation}
\end{figure}

The quantitative results of the ablation study are shown in Table~\ref{tb:ablation}.
The results show that Proposed has the best results in all evaluation methods except one, Color MSE. 
For Color MSE, w/o Perception Network performed slightly better. 
This indicates that the color of text produced by the proposed method without the Perception Network was more similar to the ground truth. 
However, as shown in Fig.~\ref{fig:ablation}, the Perception Network is required to produce reasonable results. 
In the figure, the colors are brighter without the Perception Network, but there is also a significant amount of additional noise. 
This is reflected in the results for the other evaluation measures in Table~\ref{tb:ablation}.

Also from Fig.~\ref{fig:ablation}, it can be observed how important each module is to the proposed method.
As seen in Table~\ref{tb:ablation}, the Font MSE and Color MSE are much larger for w/o Context Encoder than the proposed method. 
This is natural due to knowing the style information being provided to the network. 
There are no hints such as color, object, texture, etc.
Thus, as shown in Fig.~\ref{fig:ablation}, w/o Context Encoder only generates a basic font with no color information.
This also shows that the book cover image information is important in generating the title text.
A similar trend can be seen with w/o Discriminator and w/o Skeleton Network. 
The results show that the Discriminator does improve the quality of the font and the Skeleton Network ensures the structure of the text is robust.

\section{Conclusion\label{sec:conclusion}}

In this study, we proposed a method of generating the design of text based on context information, such as the location and surrounding image.
Specifically, we generated automatic book titles for given book covers using a neural network. The generation of the title was achieved by extracting the features of the book cover and the input text with two encoders, respectively, and using a generator with skeletal information. In addition, an adversarial loss and perception network is trained simultaneously to refine the results. As a result, we succeeded in incorporating the implicit universality of the design of the book cover into the generation of the title text. We obtained excellent results quantitatively and qualitatively and the ablation study confirmed the effectiveness of the proposed method. 
The code can be found at \url{https://github.com/Taylister/FontFits}.
In the future, we will pursue the incorporation of the text back onto the book cover.

\section*{Acknowledgement}
This work was in part supported by MEXT-Japan
(Grant No. J17H06100 and Grant No. J21K17808).

%
%
%
\bibliographystyle{splncs04}
\bibliography{icdar}
\end{document}